\documentclass{article}

\usepackage[preprint]{neurips_2021}

\usepackage[utf8]{inputenc} 
\usepackage[T1]{fontenc}    
\usepackage{url}            
\usepackage{booktabs}       
\usepackage{amsfonts}       
\usepackage{nicefrac}       
\usepackage{microtype}      
\usepackage{xcolor}         
\usepackage{graphicx}
\usepackage[rawfloats=true]{floatrow}
\usepackage[vlined,ruled]{algorithm2e}
\usepackage{xcolor}
\usepackage{subcaption}
\usepackage{siunitx}
\usepackage{array, multirow}
\usepackage{svg}
\usepackage{pythonhighlight}
\usepackage{hyperref}       
\usepackage[nocompress]{cite}

\title{Unified Distributed Environment}

\author{%
  Woong Gyu La\qquad Sunil Muralidhara \qquad Lingjie Kong \qquad Pratik Nichat \\
  \\
  Amazon Web Services \\
  \\
  \texttt{\{woong,murasuni,lingjik,pnichat\} @ amazon.com} \\
}

\SetKwProg{Fn}{Function}{}{}

\begin{document}

\maketitle

\begin{abstract}
We propose Unified Distributed Environment (UDE) \footnote{\url{https://github.com/aws-deepracer/ude}}, an environment virtualization toolkit for reinforcement learning research. 
UDE is designed to integrate environments built on any simulation platform such as Gazebo, Unity, Unreal, and OpenAI Gym. 
Through environment virtualization, UDE enables offloading the environment for execution on a remote machine while still maintaining a unified interface.
The UDE interface is designed to support multi-agent by default.
With environment virtualization and its interface design, the agent policies can be trained in multiple machines for a multi-agent environment.
Furthermore, UDE supports integration with existing major RL toolkits for researchers to leverage the benefits. 
This paper discusses the components of UDE and its design decisions.
\end{abstract}

\section{Introduction}

Recent advancements in the field of deep reinforcement learning (RL) research have led to huge breakthroughs in which AI agents can achieve performance at or better than human levels \cite{mnih2015human, schulman2017proximal, silver2017mastering, vinyals2019alphastar}. 
This leap in the research has led to the development of a wide range of tools such as OpenAI Gym \cite{brockman2016openai}, Arcade Learning Environment \cite{ale2013}, Unity ML-Agents Toolkit \cite{juliani2018unity}, Ray \cite{moritz2018ray}, and many more \cite{beattie2016deepmind, johnson2016malmo, ai2thor} to help expedite and bring more challenges to the research. 
Although these tools empower lots of RL research, the industry still lacks a fundamental tool that is cloud-friendly, and unifies the interface for environments that are built with different platforms.

RL research tasks started with simple environments such as Cart Pole, Frozen Lake, and Atari 2600 games. 
As RL research has expanded into different domain problems, new environments have been built using domain-specific platforms such as MuJoCo \cite{6386109}, PyBullet \cite{coumans2021}, VizDoom \cite{wydmuch2018vizdoom}, and Project Malmo \cite{10.5555/3061053.3061259}. 
As research advanced to tackle more realistic and practical problems, the demand for more visually and physically realistic platforms like Unity and Unreal increased \cite{rusu2018simtoreal, zhu2017target, tobin2017domain, andrychowicz2020learning, juliani2018unity}.
While these advancements helped researchers solve new problems in different domains, they also created the following drawbacks: 1) the environments built with different domain-specific platforms led to platform diversification, which added more complexity for researchers who experiment in different RL domains, and 2) the need for more realistic simulations with high-fidelity visuals and physics led to the use of heavy computational resources (CPU/GPU). This creates a bottleneck due to compute resource competition between a simulator and an RL agent's neural network policy.

In order to overcome the above two challenges, we created the Unified Distributed Environment (UDE) to support the integration of any environment built with any simulation platform. 
This is achieved by the bridge implementation for each of the platforms with a unified interface. This helps both environment developers and researchers.
The environment developers can build their environments using any platform of their choice on top of the UDE bridge SDK, making their environments compatible with UDE without any extra efforts for integration. 
Researchers can reduce their workload by using a small set of UDE interfaces instead of learning multiple independent platforms with which to integrate their work.
Ultimately, researchers can experiment against any environments built with any kind of platforms through UDE without putting any effort to make changes in their work.

Along with UDE's ability to integrate various platforms, it features an application virtualization capability to seamlessly offload the RL environments to the cloud or remote machines without any modification to its interface. 
This is extremely useful when working on realistic simulation environments, which are computationally intensive. 
With virtualization, an RL algorithm’s computation efficiency will decrease due to the competition towards the GPU/CPU resources with the simulator. 
By offloading the environment to the cloud or remote machine, both environment and RL algorithm can have their dedicated computational resources which expedites the training and evaluation processes.
Furthermore, with environment virtualization and an interface that supports multi-agent environment, UDE can distribute the agent policy trainings and rollout executions into multiple machines for further acceleration.

\section{Related Work}

In this section, we describe the related work of UDE. 
Due to recent advancements in the area of deep RL research, many toolkits have been developed to improve and expedite RL research. 
This includes providing a large set of environments and RL algorithm implementations as the baseline, and parallelizing the training to reduce the time. 
As each of the toolkits is developed to solve a particular problem, they come with their strengths and weaknesses. 
With this in mind, UDE is built to bridge the different toolkits together to re-surface each toolkit’s benefits.

\subsection{Arcade Learning Environment}

Arcade Learning Environment (ALE) \cite{ale2013} is a platform built on top of Stella, an open-source Atari 2600 emulator, to provide hundreds of Atari 2600 game environments. 
Each set of environments provides different challenges and requires different techniques to play. 
This large set of environments provides significant challenges for RL, model learning, model free learning, model-based planning, imitation learning, transfer learning, and intrinsic motivation research.

ALE has provided challenges to researchers over the past decade and has performed as a good benchmark solution. 
The environments from Atari 2600 have already been solved at or above human-level by many recent RL algorithms, so benchmarks based on ALE environments are becoming less differentiable.

\subsection{OpenAI Gym}
OpenAI Gym \cite{brockman2016openai} is a toolkit for RL research to provide a common interface for single-agent environments. It allows any environment implementation to be integrated as long as the environment supports Gym interface in Python. OpenAI Gym provides a set of simple interfaces to interact with the environment along with a set of standardized observation and action spaces to be used in different types of environments.

OpenAI Gym has allowed researchers to benchmark RL algorithms against a wide range of environments without any integration effort. This has helped researchers focus on training the agent and thereby has boosted the advancement of RL research. By integrating ALE \cite{ale2013} and Mujoco \cite{6386109} environments, OpenAI Gym provided a large set of environments to challenge researchers and allow them to bring their own custom environment.
The limitation of OpenAI Gym, however, is the support of single-agent environments only.
The support of multi-agent environments, which was listed as future work at launch, is still pending.

\subsection{Unity Engine and Unity ML-Agents Toolkit}

Unity created the Unity ML-Agents Toolkit on top of its powerful Unity Engine \cite{juliani2018unity} platform. 
The Unity game engine gives developers the capability to create more realistic simulation environments with high-fidelity visuals and physics.
The Unity ML-Agents Toolkit provides a variety of example environments as well as enables users to create custom environments with the ML-Agents SDK.
The ML-Agents SDK contains a set of RL algorithms that work with the environments out of the box and is flexible enough to allow researchers to extend it further for custom RL algorithms or build RL algorithms from scratch using \emph{ml-agents-envs}.
The SDK also supports an OpenAI Gym wrapper for experimentation with existing OpenAI Gym compatible RL algorithm implementations. 
The main restriction of the Unity ML-Agents Toolkit is that its usability is tightly coupled to environments built with Unity Engine.

\subsection{Ray and RLlib}

Ray \cite{moritz2018ray} is a general-purpose cluster-computing framework that enables distributed simulation and training for RL research. RLlib \cite{pmlr-v80-liang18b} is built on top of Ray to provide multiple RL algorithms along with OpenAI Gym environments, multi-agent environments, and support for customer environments out of the box. 
Ray and RLlib together provide end-to-end distributed RL training with minimum efforts. 

RL trainings are often sample inefficient, as it requires exhaustive exploration to find an optimal or more likely a sub-optimal policy.
This exhaustive exploration often leads to increased training time as the samples can only be collected through interaction with the environment step by step.
To improve training time, researchers often approach the process with multi-rollout strategy, which collects samples from multiple environments simultaneously.
This requires custom implementation on top of simulation environments.
With Ray, researchers can avoid the creation of custom implementation to parallelize RL training, and can also massively parallelize the RL training in actor computation.

While Ray + RLlib can be a great choice when the researchers benchmark RL environments with RL algorithms provided by RLlib, Ray + RLLib usability is at minimal during RL algorithm development as it focused on the parallelization of RL algorithm training, not the development cycle. Also, it requires researches to develop Ray + RLlib compatible algorithm implementation if they are planning to use RL algorithms outside of what RLlib offers.

\section{Unified Distributed Environment Toolkit}

\begin{figure}
  \centering
  \includegraphics[width=\textwidth]{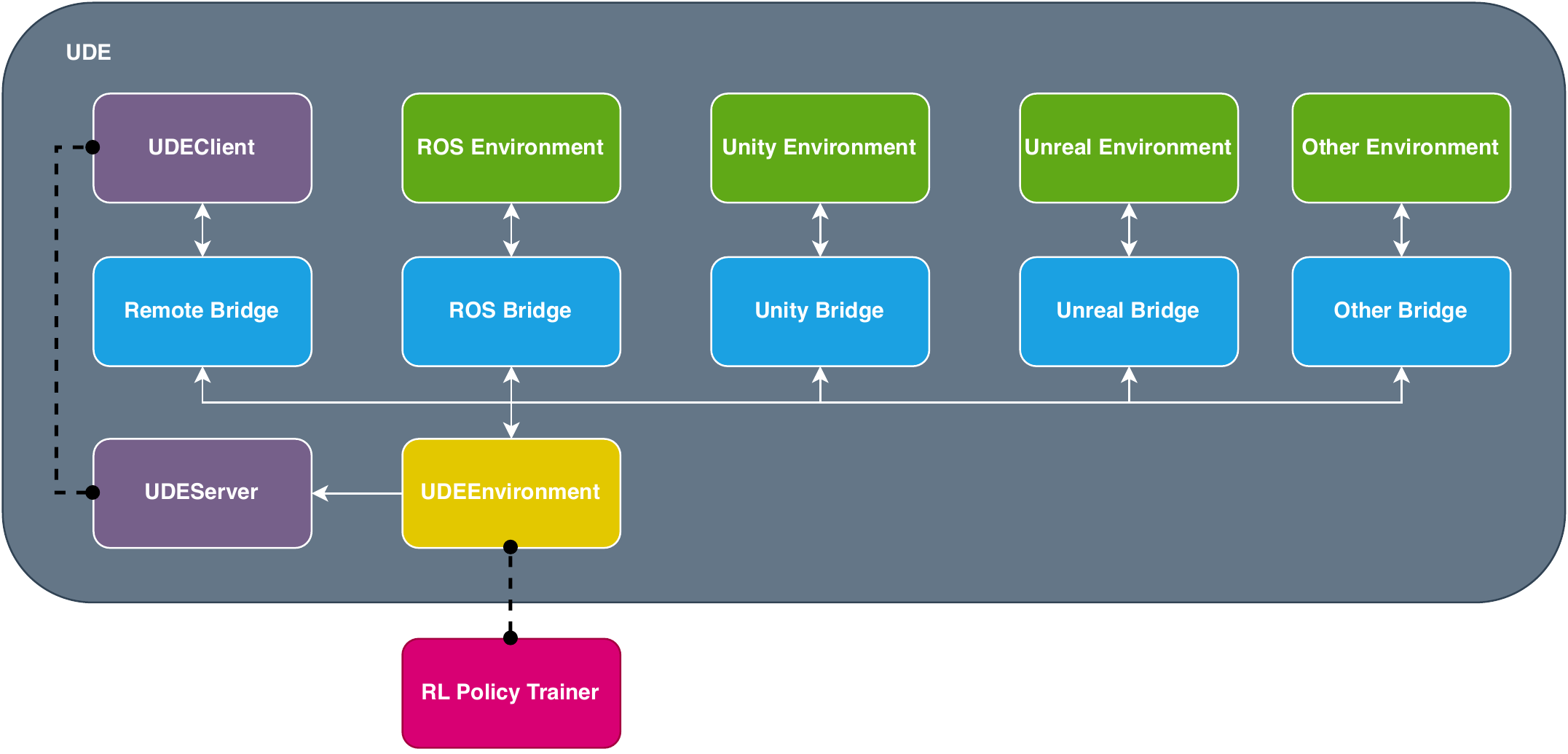}
  \caption{A Unified Distribute Environment's architecture demonstrating the uniform interface through \emph{UDEEnvironment}, and different types of bridges along with remote bridge.}
  \label{fig:ude}
\end{figure}

In this section, we describe the design of our proposed toolkit. 
UDE (Figure \ref{fig:ude}) is an environment virtualization toolkit that allows users to seamlessly offload simulation environments to the cloud or remote machines, so an agent’s neural network policy does not need to race against CPU or GPU resources with computationally heavy simulation environments. For broader simulation environment’s adaption to UDE, it is designed in a way such that any environment can be integrated to UDE toolkits through dedicated bridge SDKs.

\subsection{UDEEnvironment}

UDEEnvironment is the only interface that researchers need to work with. UDE supports the unified interface across all simulation environments, running either locally or remotely, through UDEEnvironment module. Similar to OpenAI Gym environment, UDEEnvironment has \emph{step}, \emph{reset}, and \emph{close} methods to control the environment simulation flow, and it has \emph{observation\_space} and \emph{action\_space} properties to return agents’ observation and action space information. Each UDE interface method follows similar functionalities to the OpenAI Gym interface. 
The major differences between OpenAI Gym and UDE's interfaces are that the UDEEnvironment input and output formats are in Python key-value pairs, where the key is the agent id, to support multi-agent as a default, and UDEEnvironment's \emph{step} function includes \emph{last\_action} along with \emph{observation}, \emph{reward}, \emph{done}, and \emph{info} values. These differences play an important role during multi-agent distributed training, which is explained in more detail in Section~\ref{section:remote-bridge}.

\subsection{Bridge: Platform Adaptation}

\begin{figure}
\floatsetup{footposition=caption, heightadjust=object}
\begin{floatrow}

\ffigbox{%
\begin{minipage}{0.5\textwidth}
  \centering
  \includegraphics[width=0.9\linewidth]{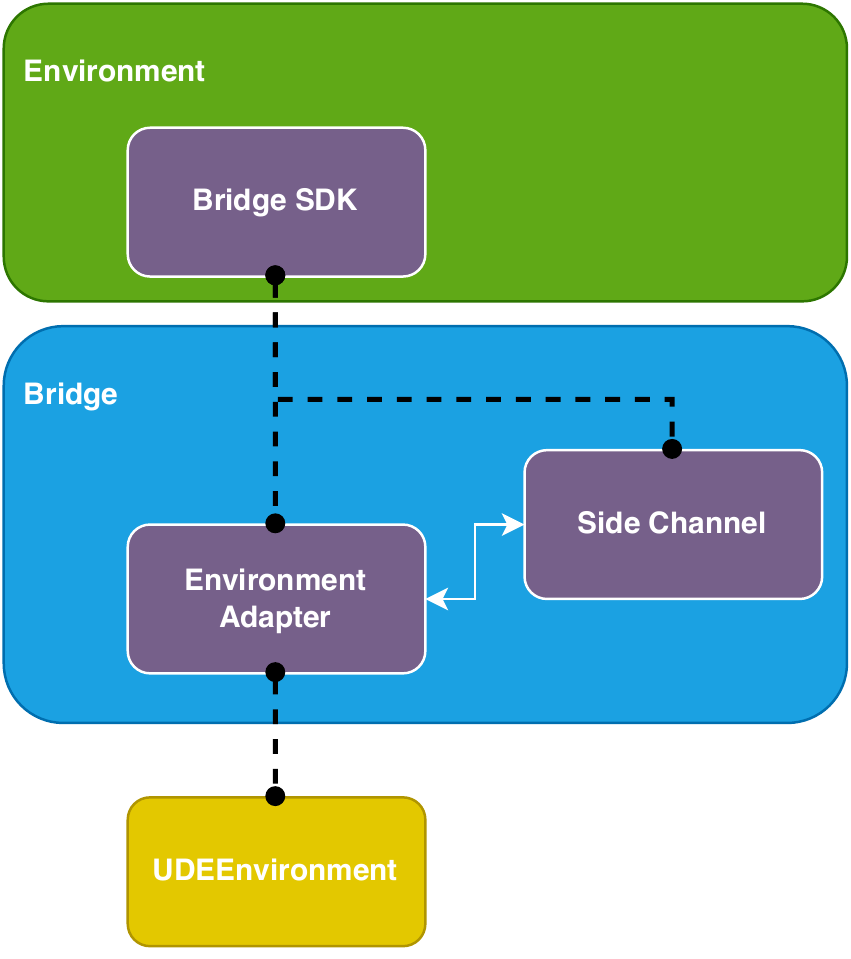}
  \caption{Bridge component architecture layout to illustrate the relationship with the actual simulation environment and the public interface.}
  \label{fig:bridge}
\end{minipage}%
}{%
}
\ffigbox{%
\begin{minipage}{0.5\textwidth}
  \centering
  \includegraphics[width=0.9\linewidth]{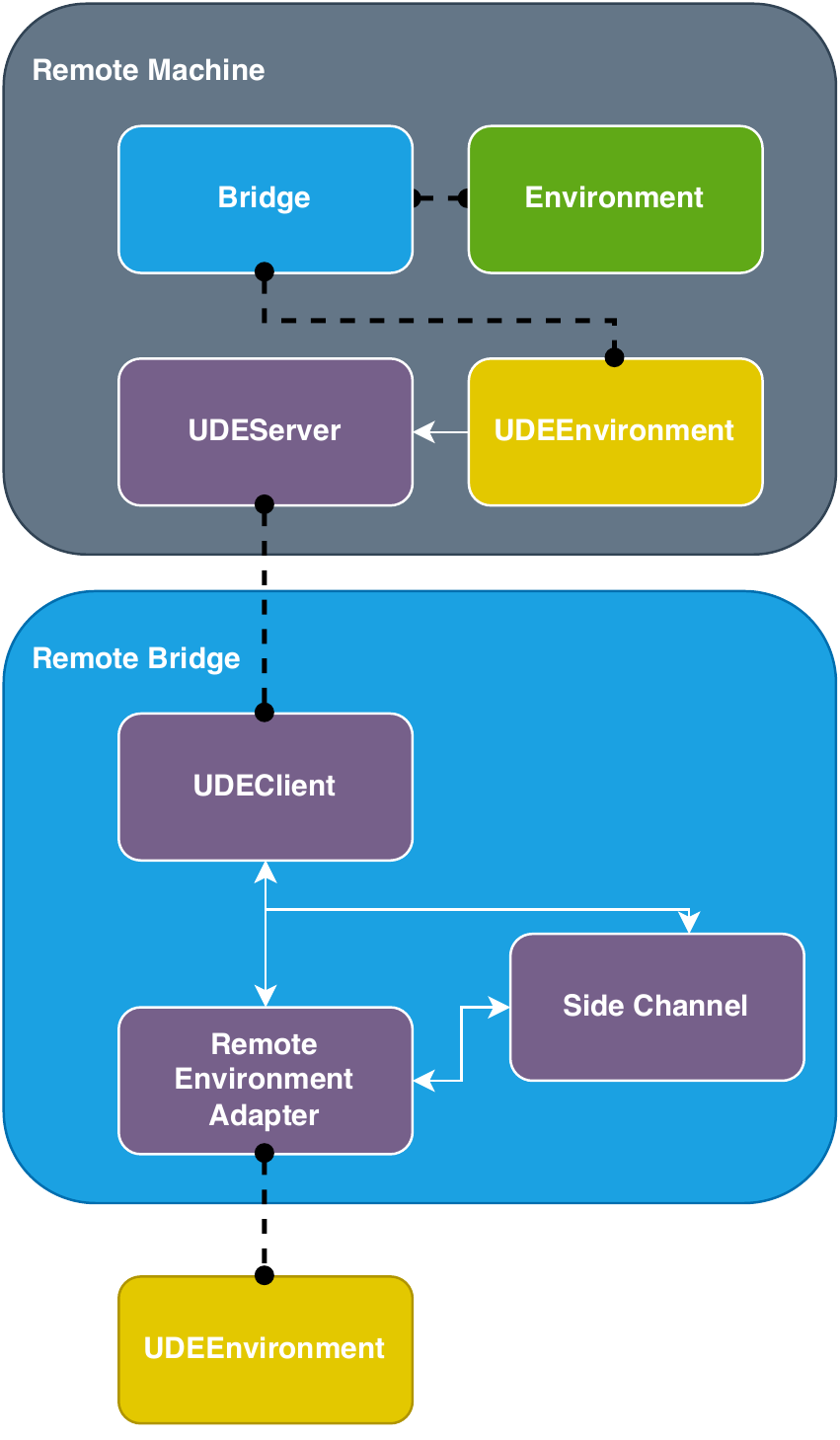}
  \caption{Remote bridge component architecture layout to illustrate the environment virtualization.}
  \label{fig:remote-bridge}
\end{minipage}%
}{%
}

\end{floatrow}
\end{figure}

As illustrated in Figure \ref{fig:bridge}, UDEEnvironment takes in \emph{environment adapter} from the environment bridge component. 
The bridge component acts as an interface for the actual simulation environment (see Appendix \ref{bridge_code} for implementation details). 
For each of the platforms, the bridge SDK is supplied to support any environments built on such platform.
Similar to Unity ML-Agents Toolkit, the UDE Bridge also contains a side-channel, which is a bi-directional communication channel to send or receive key-value pair data outside of the default environment flow control interface. 
The only difference between UDE and ML Agent's side channels is that while Unity supports arbitrarily many side-channels, UDE only supports a single side-channel for simplicity. 
This side-channel can be used for multiple purposes by dispatching it to different modules based on the key prefixes chosen by the environment developer.

\subsection{Remote Bridge: Environment Virtualization}
\label{section:remote-bridge}

\begin{algorithm}
\caption{Simplified UDEServer.\textit{Step} Procedure}
\label{alg:step}
\DontPrintSemicolon

$agentActions \leftarrow dict()$\;
$stepResultReadyEvent \leftarrow AutoResetEvent()$\;

\Fn{OnStepReceived ($agentIds$, $agentActions$)}{
 $UDEServer.Step(agentIds, agentActions)$ \;
 $stepResultReadyEvent.wait()$ \;
 \KwRet $UDEServer.stepResult$ \;
 }
 
\Fn{UDEServer.Step ($agentIds$, $agentActions$)}{
  \ForEach {$agentId, agentAction \in agentIds, agentActions$}{%
  $agentActions[agentId] \leftarrow agentAction$
  }
  
  \If {$Count(agentActions) \geq Count(agents)$}{
    $observation, reward, done, lastAction, info \leftarrow env.Step(agentActions)$\;
    $stepResult \leftarrow observation, reward, done, lastAction, info$\;
    $stepResultReadyEvent.set()$\;
    $agentActions \leftarrow dict()$\;
  }
}

\end{algorithm}

\begin{figure}
  \centering
  \includegraphics[width=0.7\textwidth]{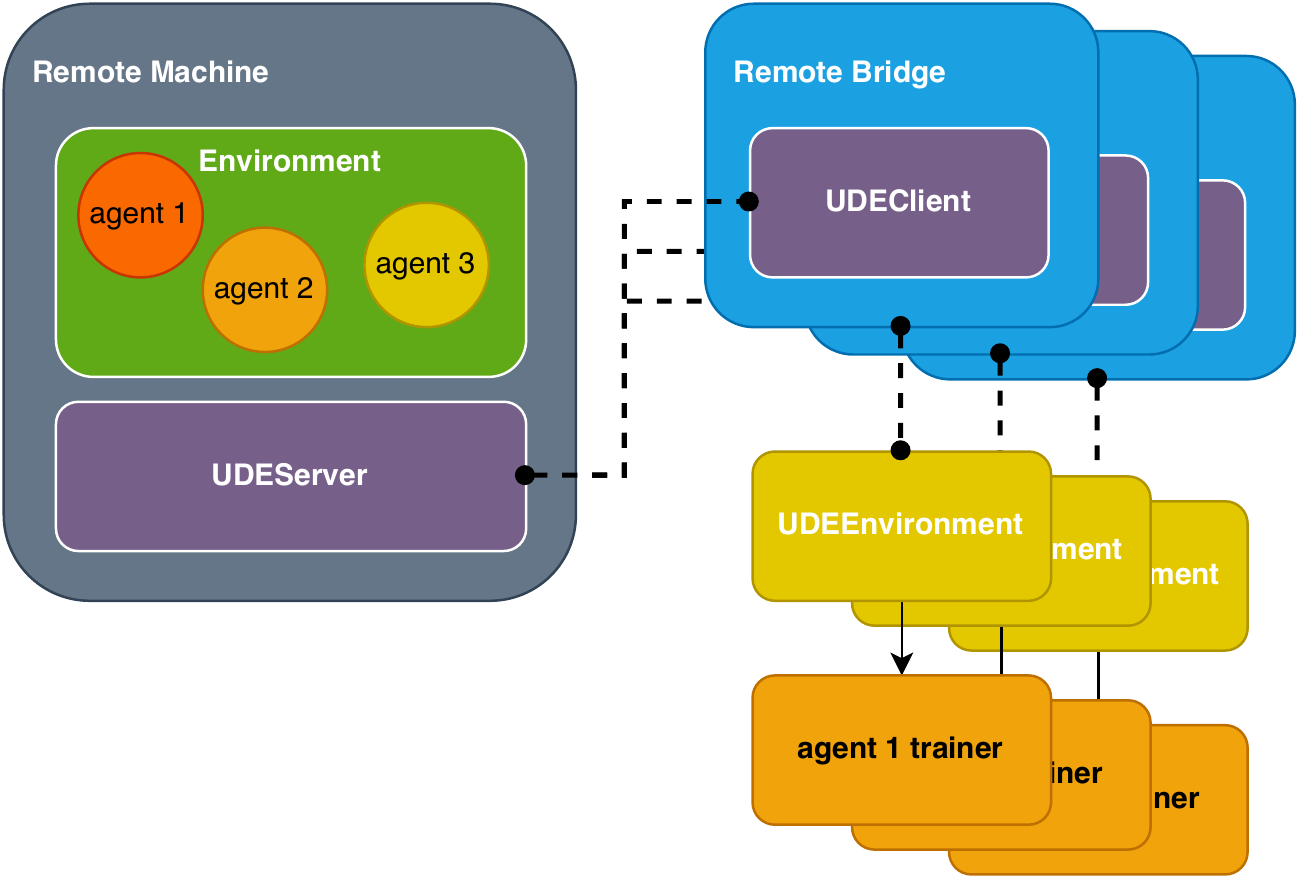}
  \caption{Illustration of multiple UDEClient connections to an UDEServer servicing multi-agent environment.}
  \label{fig:multi-agent}
\end{figure}

One of the key components for environment virtualization is the remote bridge.
As illustrated in Figure \ref{fig:remote-bridge}, unlike other UDE bridges, remote bridge contains UDEClient which can communicate with the UDEServer via a gRPC communication protocol. 
The UDEServer takes in UDEEnvironment with the actual simulation environment and services the UDEEnvironment feature to connected UDEClients  (see Appendix \ref{remote_code} for the remote bridge usage). One of the advantages of environment virtualization is that UDE supports distributed agent policy training for multi-agent environments. As we can see in Figure \ref{fig:multi-agent}, UDEServer allows the connections from multiple UDEClients where each client can represent one or more agents in the environment. With this capability, UDE supports the distribution of multi-agent policy training processes on multiple machines. 

Algorithm \ref{alg:step} illustrates the procedure of UDEServer's environment \emph{step} execution handling. 
The gRPC server invokes the \emph{OnStepReceived} function, whenever the \emph{step} execution is requested by UDEClient with agent action(s). UDEServer simply delays until all of the agents’ actions are received, and executes a single step function call to the actual simulation environment. 
All of the agents’ observations, rewards, last actions, done flags, and environment information will be shared to every connected UDEClient. Hence, even if agent policies reside in different machines, they can be trained with full visibility of other agents' information.

\subsection{OpenAI Gym Integration}
\label{openai_bridge}
UDE allows two integration points with the OpenAI Gym to broaden the adaptability. 
First, UDE supports OpenAI Gym environments through OpenAI Gym bridge SDK. 
The OpenAI Gym bridge SDK also allows researchers to change the OpenAI Gym environment through a side-channel dynamically. Once researchers implement algorithms compatible to UDE, they can experiment with all OpenAI Gym environments along with other environments built with different platforms without any change in their algorithm implementations.
 
Second, UDE also provides an OpenAI Gym wrapper module, which wraps UDEEnvironment and interfaces as the OpenAI Gym environment interface (see Appendix \ref{gym_wrapper_code} for the details). Using this module, all UDE compatible simulation environments can be interfaced as the OpenAI Gym environment interface. 
The only drawback is that the interface's limited support of single-agent-only environments carries over as well.
This allows researchers to re-use their algorithm implementations for OpenAI Gym environments with UDE and still leverage UDE’s virtualization and adaptation feature.

\subsection{Ray Integration}

Ray’s primary focus is parallelizing the processes between CPUs, GPUs, and clusters, so Ray mainly distributes the RL training by starting multiple simulation environments and uses multiple rollout workers to collect the samples in parallel, and, then, the agent's neural network policy can also be trained in parallel by the multi-GPU plan. UDE’s focus is not parallelizing the RL training but virtualizing the simulation environments by abstraction. 
To allow the researchers to leverage the advantages of both Ray and and UDE, UDE can be used with Ray as a custom Ray environment and supply Ray with wider adaptation of simulation environments in different platforms where researchers can benefit from expedited benchmarking through Ray's distributed training.

\section{Experiments}

We evaluated the agent's performance of three different environment settings using UDE and compared the performances of those evaluated against the native \textbf{OpenAI Gym environment} setting. 
First, the \textbf{env using UDE} setting is where both the simulation environment and training algorithm running on the same process and simulation environment interfaced through UDE.
Second, the \textbf{remote env in local host} setting is where the simulation environment is executed on a process different from the training algorithm using UDE's environment virtualization. Here both processes reside on the same host machine.
Lastly, the \textbf{env from remote host} setting is where the simulation environment and training algorithm are running on different host machines and communicated over the network using UDE's environment virtualization.

\textsc{Pendulum}, \textsc{Lunar Lander Continuous}, and \textsc{Hopper} environments from the OpenAI Gym were used for the experiments to represent environments with different difficulty levels --- easy, medium, and hard respectively.
These OpenAI Gym environments are interfaced through UDE using the OpenAI Gym bridge SDK described in Section \ref{openai_bridge}.
We trained the policies using  DDPG\cite{lillicrap2015continuous}, PPO\cite{schulman2017proximal}, and SAC\cite{haarnoja2018soft} algorithm implementations.
For each algorithm, we trained five different instances with different random seeds and performed five evaluation rollouts every 1,000 environment steps.
The hyper-parameters used for each algorithm training are presented in Appendix \ref{hyperparameters}. 

In Figure \ref{fig:Pendulum_Hopper_LunarLander}, we present the total average return of evaluation rollouts for each of the algorithms --- DDPG, PPO, and SAC --- we trained in the \textsc{Pendulum}, \textsc{Lunar Lander Continuous}, and \textsc{Hopper} environments.
The solid curves correspond to the mean and the shaded regions correspond to the standard deviation of the returns over the five trials.
Regardless of simulation environments and training algorithms, each UDE environment setting converges in the same trend as the reference, the native OpenAI Gym environment setting. 
This shows that the use of UDE has no negative impact to the agent's performance for different types of RL policy training. 
Particularly, UDE allows researchers to safely offload the environment to be executed remotely without risking the agent's performance with the additional benefit of speeding up the training by running multiple environments on different hosts.

\begin{figure}
    \centering
    \begin{subfigure}{0.32\textwidth}
        \centering
        \includegraphics[width=\textwidth]{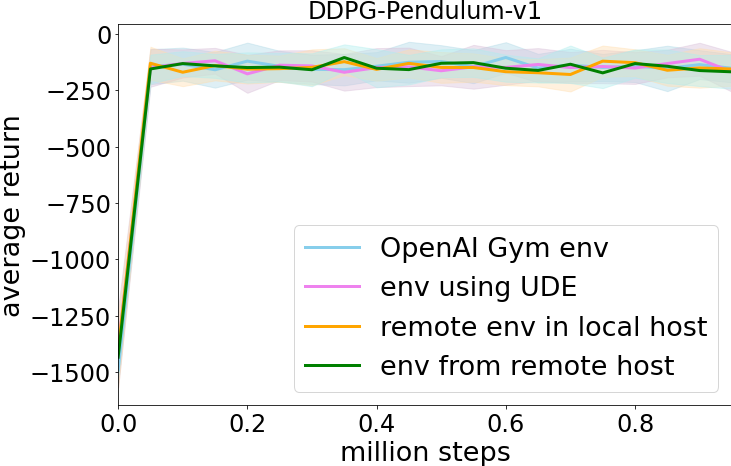}
        \label{fig:ddpg_pendulum}
    \end{subfigure}
    \begin{subfigure}{0.32\textwidth}
        \centering
        \includegraphics[width=\textwidth]{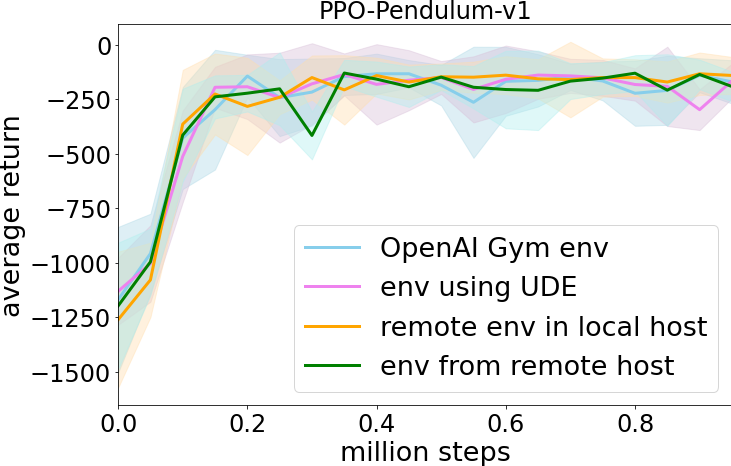}
        \label{fig:ppo_pendulum}
    \end{subfigure}
    \begin{subfigure}{0.32\textwidth}
        \centering
        \includegraphics[width=\textwidth]{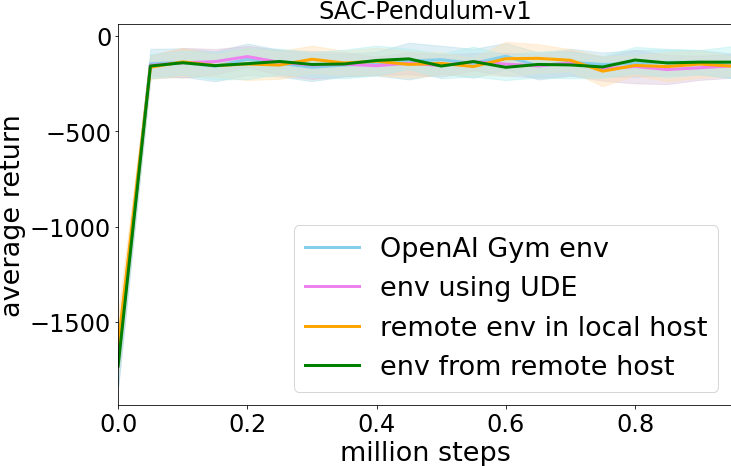}
        \label{fig:sac_pendulum}
    \end{subfigure}
    \begin{subfigure}{0.32\textwidth}
        \centering
        \includegraphics[width=\textwidth]{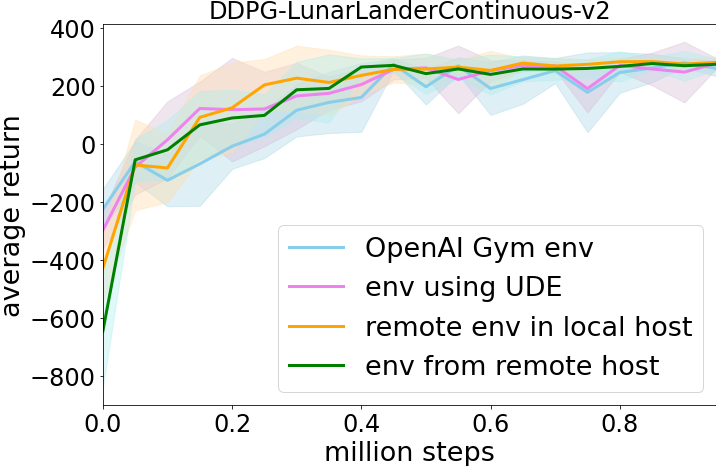}
        \label{fig:ddpg_lunar}
    \end{subfigure}
    \begin{subfigure}{0.32\textwidth}
        \centering
        \includegraphics[width=\textwidth]{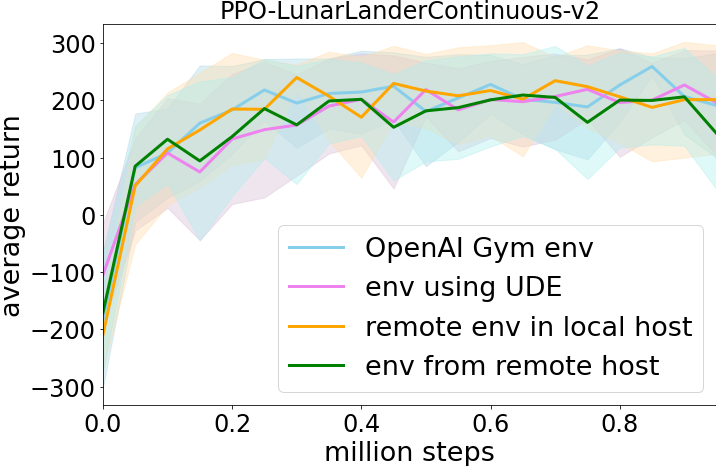}
        \label{fig:ppo_lunar}
    \end{subfigure}
    \begin{subfigure}{0.32\textwidth}
        \centering
        \includegraphics[width=\textwidth]{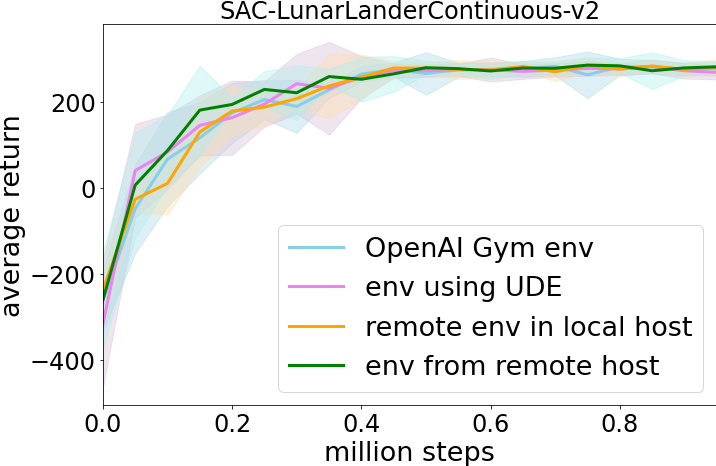}
        \label{fig:sac_lunar}
    \end{subfigure}
    \begin{subfigure}{0.32\textwidth}
        \centering
        \includegraphics[width=\textwidth]{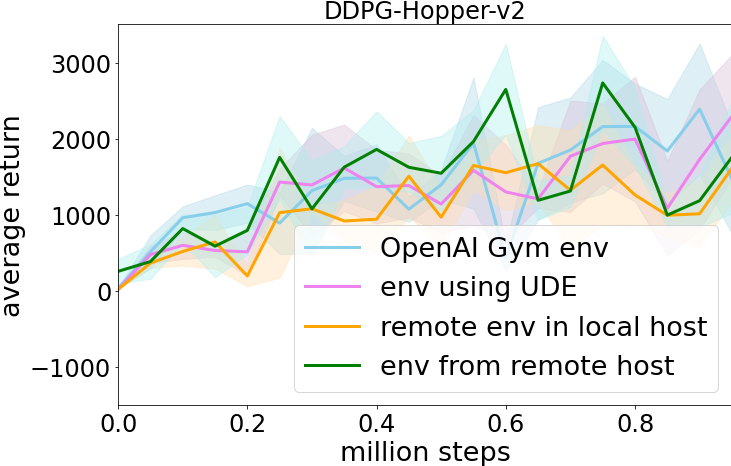}
        \label{fig:ddpg_hopper}
    \end{subfigure}
    \begin{subfigure}{0.32\textwidth}
        \centering
        \includegraphics[width=\textwidth]{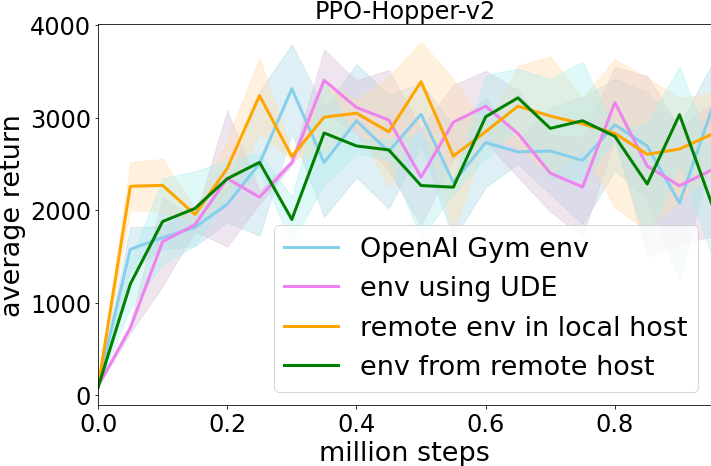}
        \label{fig:ppo_hopper}
    \end{subfigure}
    \begin{subfigure}{0.32\textwidth}
        \centering
        \includegraphics[width=\textwidth]{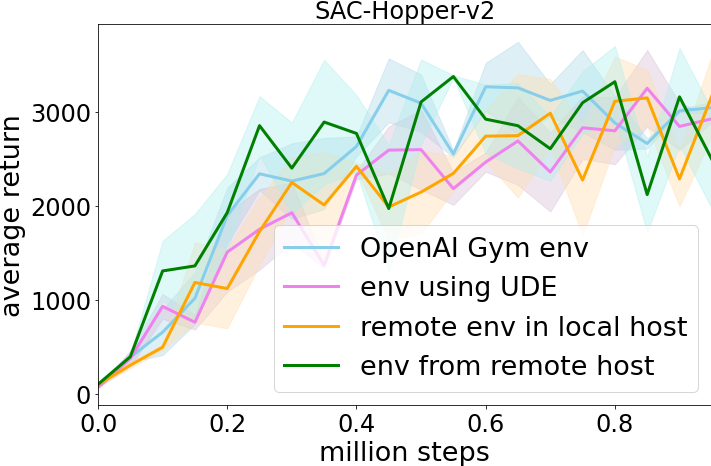}
        \label{fig:sac_hopper}
    \end{subfigure}
    \caption{The agent performance in Pendulum, Lunar Lander Continuous, and Hopper environments using DDPG, PPO, and SAC algorithms with different UDE settings.}
    \label{fig:Pendulum_Hopper_LunarLander}
\end{figure}

\section{Conclusion}

In this paper, we have presented the design of the Unified Distributed Environment, a toolkit for virtualizing the RL environment. 
UDE is not meant to replace any of the existing tools or toolkits that have been developed to support RL algorithm research, but, rather, bridge them together to synthesize the research environment and further boost research by providing the researcher with a light-weight interactive development environment through offloading the heavy lifting of environment simulation to the cloud or a remote machine. Currently, UDE supports ROS and OpenAI Gym bridges with expanded support for Unity and Unreal planned.

\bibliography{main}
\bibliographystyle{unsrt}

\newpage
\appendix








\section{Platform Adaptation}
\label{bridge_code}

\subsection{OpenAI Gym Local UDEEnvironment}
\begin{python}
from ude_gym_bridge import GymEnvironmentAdapter
from ude import UDEEnvironment

# initiate an OpenAI Gym environment adapter.
adapter = GymEnvironmentAdapter(env_name="CartPole-v0")
env = UDEEnvironment(ude_env_adapter=adapter)

# reset the environment and retrieve first observation of each agent 
# in dictionary format (ex. agent_id: observation) along with
# environment information. 
state, info = env.reset() 
for _ in range(100):
    # sample a random action from agent0
    action = env.action_space["agent0"].sample()
    # submit agent's action and retrieve observation, reward, done, 
    # and action of each agent along with environment information.
    state, reward, done, action, info = env.step({"agent0": action})
env.close()

\end{python}
\subsection{ROS Local UDEEnvironment}
\begin{python}
from ude_ros_bridge import ROSEnvironmentAdapter
from ude import UDEEnvironment

# initiate a ROS environment adapter.
adapter = ROSEnvironmentAdapter()
env = UDEEnvironment(ude_env_adapter=adapter)
...
\end{python}

\section{Environment Virtualization}
\label{remote_code}
The environment can be provided as a service using \emph{UDEServer}.
To access this environment, \emph{RemoteEnvironmentAdapter} should be instantiated with the hostname and the port of the remote environment. This \emph{RemoteEnvironmentAdapter} instance is used as an argument to create \emph{UDEEnvironment} instance.
Through \emph{UDEEnvironment} instance, users can interact with remote environment as local environment.

\begin{python}
# On a remote machine:
# adapter = EnvironmentAdapter()
# env = UDEEnvironment(ude_env_adapter=adapter)
# server = UDEServer(ude_env=env)
# server.start()
# ...

from ude import RemoteEnvironmentAdapter, UDEEnvironment

# initiate remote environment adapter with hostname and port
adapter = RemoteEnvironmentAdapter(address=HOSTNAME, port=PORT)
env = UDEEnvironment(ude_env_adapter=adapter)
...
\end{python}

\newpage
\section{UDEEnvironment as OpenAI Gym Environment}
\label{gym_wrapper_code}

The \emph{UDEEnvironment} can be interfaced as OpenAI Gym Environment using \emph{UDEToGymWrapper} class as shown in example below. The example is demonstrated with \emph{RemoteEnvironmentAdapter} use-case, but this works with any adapters (ex. \emph{GymEnvironmentAdapter}, \emph{ROSEnvironmentAdapter}, \emph{UnityEnvironmentAdapter}, etc.).
\begin{python}
from ude import RemoteEnvironmentAdapter, UDEEnvironment, UDEToGymWrapper

# initiate remote environment adapter with hostname and port
adapter = RemoteEnvironmentAdapter(address=HOSTNAME, port=PORT)
env = UDEEnvironment(ude_env_adapter=adapter)
env = UDEToGymWrapper(ude_env=env)

# reset the environment and retrieve first observation
state = env.reset() 
for _ in range(100):
    # sample a random action
    action = env.action_space.sample()
    # submit agent's action and retrieve observation, reward, and done 
    # of the agent along with environment information.
    state, reward, done, info = env.step(action)
env.close()
\end{python}

\section{Hyperparameters}
\label{hyperparameters}
\begin{center}
\begin{tabular}{||r || r | l |} 
 \hline
 \multirow{8}{*}{\textbf{DDPG}} & learning rate $\alpha$ & \num{1e-3} \\\cline{2-3}
        &buffer size & 1,000,000 \\\cline{2-3}
        &learning starts & 100 steps \\\cline{2-3}
        &batch size & 100 \\\cline{2-3}
        &soft update coefficient $\tau$ & 0.005 \\\cline{2-3}
        &discount factor $\gamma$ & 0.99 \\\cline{2-3}
        &training frequency & 1 episode \\\cline{2-3}
        &gradient steps & \# of rollout steps \\
 \hline\hline
 \multirow{8}{*}{\textbf{PPO}} & learning rate $\alpha$ & \num{3e-4} \\\cline{2-3}
        &batch size & 64 \\\cline{2-3}
        &\# of epoch & 10 \\\cline{2-3}
        &discount factor $\gamma$ & 0.99 \\\cline{2-3}
        &gae lambda $\lambda$ & 0.95 \\\cline{2-3}
        &clipping range & 0.2 \\\cline{2-3}
        &entropy coefficient & 0.0 \\\cline{2-3}
        &value function coefficient & 0.5 \\
 \hline\hline
 \multirow{9}{*}{\textbf{SAC}} & learning rate $\alpha$ & \num{3e-4} \\\cline{2-3}
        &buffer size & 1,000,000 \\\cline{2-3}
        &learning starts & 100 steps \\\cline{2-3}
        &batch size & 256 \\\cline{2-3}
        &soft update coefficient $\tau$ & 0.005 \\\cline{2-3}
        &discount factor $\gamma$ & 0.99 \\\cline{2-3}
        &training frequency & 1 step \\\cline{2-3}
        &gradient steps & 1 \\\cline{2-3}
        &target update interval & 1 rollout \\
 \hline\hline   
 \end{tabular}
\end{center}

\end{document}